\pdfoutput=1
\documentclass{article}

\PassOptionsToPackage{square,sort,comma,numbers,compress}{natbib}
\usepackage[final]{neurips_2020}

\usepackage[utf8]{inputenc} 
\usepackage[T1]{fontenc}    
\usepackage{hyperref}       
\usepackage{url}            
\usepackage{booktabs}       
\usepackage{amsfonts}       
\usepackage{nicefrac}       
\usepackage{microtype}      

\usepackage[linewidth=1pt]{mdframed}
\usepackage{tikz-dependency}

\usepackage{graphicx}
\usepackage{color}
\usepackage{wrapfig}
\usepackage{listings}
\usepackage{graphbox}
\usepackage{multirow}
\NewEnviron{elaboration}{
\par
\begin{tikzpicture}
\node[rectangle,minimum width=\textwidth] (m) {\begin{minipage}{.98\textwidth}\BODY\end{minipage}};
\draw[dashed] (m.south west) rectangle (m.north east);
\end{tikzpicture}
}
\usepackage{enumitem}

\usepackage[export]{adjustbox}

\usepackage{scalerel,xparse}

\title{Playing Text-Based Games with Common Sense}

\author{%
  Sahith Dambekodi\thanks{Equal contribution; authors listed alphabetically.}, ~Spencer Frazier$^{*}$, Prithviraj Ammanabrolu, and Mark O. Riedl \\
  Georgia Institute of Technology\\
  \texttt{\{sdambekodi3, sfrazier7, raj.ammanabrolu, riedl\}@gatech.edu} \\
}

\begin{document}

\maketitle
\graphicspath{ {graphs/} }
\begin{abstract}

Text-based games are simulations in which an agent interacts with the world purely through natural language.
They typically consist of a number of puzzles interspersed with interactions with common everyday objects and locations. 
Deep reinforcement learning agents can learn to solve these puzzles.
However, the everyday interactions with the environment, while trivial for human players, present as additional puzzles to agents. 
We explore two techniques for incorporating commonsense knowledge into agents.
(1)~ Inferring possibly hidden aspects of the world state with either a commonsense inference model (COMET~\citep{Bosselut2019COMETCT}), or a language model (BERT~\citep{devlin18}).
(2)~Biasing an agent's exploration according to common patterns recognized by a language model. 
We test our technique in the {\em 9:05} 
game, which is an extreme version of a text-based game that requires numerous interactions with common, everyday objects in common, everyday scenarios.
We conclude 
that agents that augment their beliefs about the world state with commonsense inferences are more robust to observational errors and omissions of common elements from text descriptions.

\end{abstract}

\section{Common Sense in Text-Based Worlds}

\begin{figure}[t]
\centering
\begin{minipage}{0.47\linewidth}
\scriptsize
\begin{mdframed}
\begin{elaboration}
  \parbox{.85\textwidth}{
\emph{Observation:} This bedroom is extremely spare, with dirty laundry scattered haphazardly allover the floor. Cleaner clothing can be found in the dresser. A bathroom lies to the south,while a door to the east leads to the living room. On the end table are a telephone, a wallet and some keys.The phone rings.
}
\end{elaboration}
\begin{flushleft}
\emph{Action:} \textbf{Answer phone}
\end{flushleft}
\begin{elaboration}
  \noindent\parbox{.98\textwidth}{
\emph{Observation:} You pick up the phone. ``Hadley!'' a shrill voice cries. ``Hadley, haven’t you even left yet?? You knew that our presentation was at nine o’clock sharp! First the thing with the printers, now this - there won’t even be enough left of you for Bowman to fire once he's done with you. Now get the hell down here!!''
}
\end{elaboration}

\begin{flushleft}
\emph{Action:} \textbf{Examine me}
\end{flushleft}
\begin{elaboration}
  \noindent\parbox{.99\textwidth}{
\emph{Observation:} You're  covered  with  mud  and  dried  sweat.   It  was  an  exhausting  night  - no wonder you overslept!   Even making it to the bed before conking out was a heroic accomplishment.
}
\end{elaboration}
\end{mdframed}
\end{minipage}
~
\includegraphics[align=c,width=0.5\linewidth]{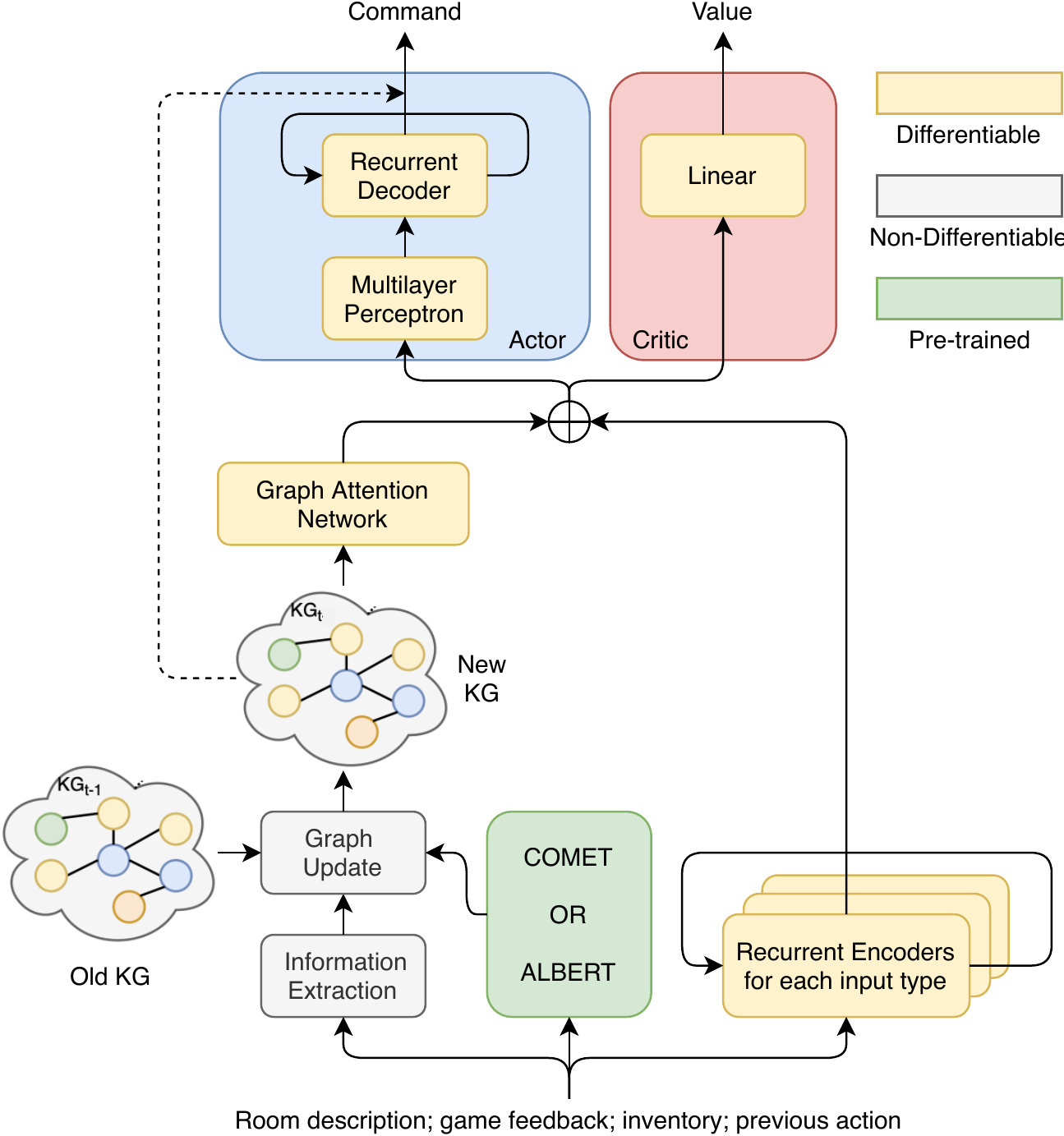}
\caption{Excerpt from the {\em 9:05} text-based game (left).
KG-A2C with augmentations (right).}
\label{fig:zorkexcerpt}
\end{figure}

\textit{Text-based games}---also called interactive fictions or text adventures---are simulations in which an agent interacts with the world purely through natural language by reading textual descriptions of the current state of the environment and composing text commands to enact change in the environment.
Text-based games are partially observable in the sense that the agent can only observe the details of one particular ``room'' or location at a time. 
See Figure~\ref{fig:zorkexcerpt} (left) for a sample text-based game interaction.
Text-based games can have a very high branching factor; the popular commercial text-adventure, {\em Zork} has $1.64 \times 10^{14}$ possible action commands that can be entered per state~\citep{jericho}.
A number of agents have been developed to attempt to play text-based games~\citep{narasimhan15,fulda17,zahavy18,ammanabrolu,adolphs2019ledeepchef,jonmay,jericho,Ammanabrolu2020Graph,Ammanabrolu2020HowTA,Murugesan2020EnhancingTR}.
State of the art text-based game playing agents use {\em knowledge graphs} to represent the agent's belief about the full world state of the game.
A knowledge graph is a set of $\langle subject,relation,object\rangle$ triples.
The knowledge graph is encoded into a deep reinforcement learning agent that learns to infer the best action to take based on this state information.
KG-A2C~\citep{Ammanabrolu2020Graph}, in particular, uses information extraction techniques and rules to identify nodes and relations from text descriptions. 

Many real-world activities can be thought of as a sequence of sub-goals in a partially observable environment.
These activities---getting ready to go to work, for example---are considered trivial for humans because of {\em commonsense knowledge}.
Commonsense knowledge is a set of facts, beliefs, and procedures shared among many people in the same society or culture. 
However, to an agent that knows only what it has learned from interacting with the environment, even tasks that humans take for granted can involve considerable trial-and-error effort.
In this paper, we look at how commonsense knowledge can be incorporated in to text-based reinforcement learning agents that operates in games that simulate common, everyday situations. 
We hypothesize that access to commonsense knowledge can enable an agent to more quickly converge on a policy that completes common, everyday tasks.
We further hypothesize that commonsense knowledge can allow the agent to infer the presence of elements in the world when observations are noisy or fail.
For example, if the agent fails to observe the presence of a sink in a bathroom, it can still infer that an action such as \texttt{wash hands} may still be applicable given common knowledge of what a bathroom contains. 

We experiment with agents in the text-game, {\em 9:05},\footnote{{\em 9:05} is part of the {\em Jericho}~\citep{jericho} suite of games at \url{https://github.com/microsoft/jericho}} 
which is a ``slice of life'' simulation game.
The player must get ready for work, navigate a simple drive to work, and perform some workplace interactions.
It is one of the hardest games for an agent to solve~\citep{jericho}; to date no agent with unrestricted ability to generate commands has completed the game.
Other text-based game playing techniques that incorporate commonsense knowledge include using semantic word vectors to infer actions that can be applied to objects~\cite{fulda17}, or
looking up information about objects in ConceptNet~\citep{Murugesan2020EnhancingTR}.

We experiment with ways to incorporate commonsense knowledge into a deep reinforcement learning game playing agent.
The first approach is to incorporate commonsense knowledge into the world state, specifically the knowledge graph, which is the agent's beliefs about the world state.
We look at two sources of commonsense inference.
COMET~\citep{Bosselut2019COMETCT} is a neural model that takes a simple sentence and infers what will be commonly believed about the people and objects referenced in the sentence.
An alternative source of commonsense knowledge is BERT~\citep{devlin18}, which is believed to have acquired significant commonsense knowledge from texts on which it was trained~\citep{Zhou2020EvaluatingCI}.
Commonsense knowledge also manifests itself as procedural knowledge---common situations are addressed by following familiar patterns of behavior.
Our third technique is to bias the agent toward certain sequences of action commands using BERTs next sentence prediction mode.



\section{Methods}

We experiment with three agent designs, each using a different source or a different way of incorporating commonsense knowledge into the agent.
All three agents build off the KG-A2C~\citep{Ammanabrolu2020Graph} agent framework, which is shown in Figure~\ref{fig:zorkexcerpt} (right).
At every step, KG-A2C uses an information extraction process to identify $\langle subject, relation, object\rangle$ triples in text descriptions of the current location. 
These triples are added to an ever-growing knowledge graph, which is encoded into the neural architecture of the agent to inform the choice of command. 
The knowledge graph is the agent's belief about the state of the world.
Commands that reference objects are filtered out if they reference entities not in the knowledge graph.
KG-A2C is our baseline.

\textbf{The Q*BERT Agent.}
The first agent is Q*BERT~\citep{Ammanabrolu2020HowTA}, which augments the knowledge graph by using
the pre-trained language model, ALBERT~\citep{Lan2020ALBERT:},
a variant of BERT
that is fine-tuned for question answering.
Q*BERT generates questions about the current environment and ALBERT proposes answers, which are converted into $\langle subject, relation, object\rangle$ and added to the knowledge graph (see Figure~\ref{fig:zorkexcerpt} (right)).
While Q*BERT was not explicitly design with commonsense knowledge in mind, we hypothesize that ALBERT is linking text observations to a broader set of knowledge about the world that has been acquired through training on a very large corpus of texts.
That is, Q*BERT may infer things about the environment that are not directly observed but are also commonly believed by humans in similar situations.

\textbf{The COMET-A2C Agent}.
Our second agent is similar to Q*BERT but replaces ALBERT with COMET~\cite{Bosselut2019COMETCT}, a neural commonsense inference model. 
Unlike ALBERT, COMET was trained on the ConceptNet~\cite{Speer2016ConceptNet} dataset to take text sentences and generate a number of short phrases that people are likely to directly infer.
COMET produces several types of inferences.
We use COMET's {\em HasA} inference class. 
COMET-A2C uses KG-A2C's information extraction process to produce $\langle subject,relation,object\rangle$ triples, and then we add additional  nodes inferred by COMET. 
We hypothesize that COMET will make the agent's understanding of the world state more robust by adding object commonly found in certain types of locations.
See Figure~\ref{fig:zorkexcerpt} (right) which shows the addition of COMET or ALBERT to the KGA2C model. The updated knowledge graph is sent as input to the Graph Attention Network which converts nodes into features and applies self-attention on all these features. The output of the model is then embedded with the encoded vector outputs for the state feedback of the environment. This final embedded vector is sent to the Actor-Critic model to decode the action.

\textbf{KG-A2C-BERT.}
Our third agent is identical to KG-A2C, except that it uses a policy-shaping approach to exploration~\cite{griffith2013policy}.
KG-A2C-BERT samples the top $k$ commands generated by the network and scores each based on a history of previous commands.
This is done by concatenating the currently proposed command to prior commands and computing the likelihood of that entire text sequence using BERT to compute $Pr(c_t | c_1...c_{t-1}; \theta)$ where $c_i$ is a command at time step $t$ and $\theta$ is BERT's pre-trained weights.
The $k$ candidate commands are re-ranked according to the score and the agent re-samples from the new distribution.
In this way, KG-A2C-BERT is biased toward exploring commands which logically entail one another, according to what BERT has learned from NSP pretraining on its very large corpus of text.
In Figure~\ref{fig:zorkexcerpt} (right), the green box is removed.

\textbf{Environment.}
We conduct experiments in the {\em 9:05} slice of life text-based game. 
In this game, the player must get ready for work by taking a shower, wearing clean clothes and then travel to the workplace by car. There are a large variety of different tasks that can be performed by the agent, most of which have no impact on reaching the end goal. There are very few sequences that reach the end of the game; and each of these sequences are of 25 to 30 specific actions. 
The {\em 9:05} game provides a single score at the end of the game: a score of 1 at the end or 0 if the player fails.
Due to the extreme sparseness of feedback all agents struggle to make any significant progress except by accident; the branching factor is too high and the only reward feedback requires 25-30 steps executed in the perfect order.
Consequently, we provide a shaped reward function. 
We define a sequence of observations that the agent should see if progress is being made and give +1 reward for each observation: 
entering the bedroom, entering the bathroom, taking off the watch, taking off soiled clothes, dropping clothes. and entering the shower. The critical checkpoints are at reward 2 for when the agent enters the bathroom and at reward 6 for when the agent successfully enters the shower.
The reward states are chosen such that they can only be observed in one particular order and that loops cannot occur.

\textbf{Experiments.}
We conducted two experiments.
(1)~In the first experiment, we used the version of {\em 9:05} where reward is given for passing key states. 
(2)~In the second experiment, we test agents' abilities to 
supplement missing/failed observations with commonsense inferences. 
We introduce a modified version of {\em 9:05} that has the shaped reward but also deletes all textual references to \texttt{sink}, \texttt{toilet}, and \texttt{shower} from the description of the \texttt{bathroom} location.
This simulates the situation in which the agent's observations have failed to notice any of the objects, or to correctly parse and extract relations pertaining to these objects. 
It also simulates the situation in which a human deems it unnecessary or obvious to state the presence of these objects in a bathroom.
The player must interact with all three objects in order to progress in the game.
The player can still interact with the objects even they are not present in the description text; objects were not removed, only the text mentions.

\section{Results and Discussion}

\begin{figure*}[t]
\includegraphics[width=0.5\linewidth]{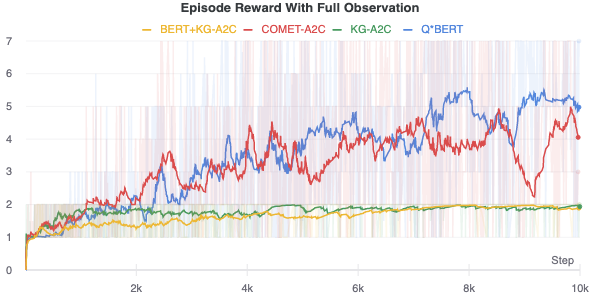}
\includegraphics[width=0.5\linewidth]{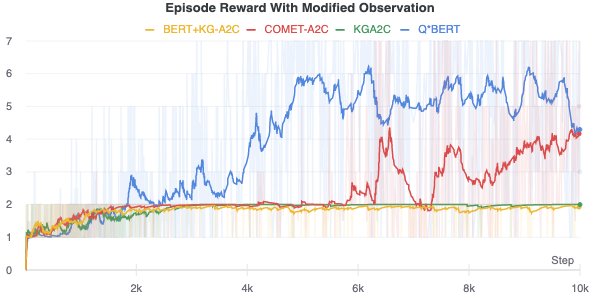}
\caption{Reward performance for all agents on {\em 9:05} with full observations (left) or modified observations (right). 
The solid lines show smoothed average performance over 4 runs, with faded bars showing max across runs.
A reward of 2 indicates that the agent has entered the bathroom and a reward of 6 indicates that the agent has entered the shower. 
}
\label{fig:graphs}
\centering
\end{figure*}

The results of Experiment 1 are shown in Figure~\ref{fig:graphs} (left).
KG-A2C never makes it to the shower and gets stuck after it enters the bathroom.
KG-A2C-BERT sometimes makes it to reward 5 (dropping clothes just before getting in the shower), but not reliably so its average performance is similar to that of KG-A2C.
COMET-A2C and Q*BERT are both able to reliably get past the shower and on to the next phase of the game where the player must drive to work. 
Their performances in this experiment are not significantly different.
%
This experiment tells us that the commonsense knowledge helps agent performance in {\em 9:05}, which makes heavy use of locations and situations that also commonly occur in the real world.
KG-A2C-BERT performs better than KG-A2C because BERT informs the agent's exploration by comparing action command sequences to patterns BERT recognizes.
COMET-A2C adds {\em HasA} relations to the knowledge graph; its improvement over baseline and KG-A2C-BERT is due to inferring entities that might not have been properly extracted from text descriptions.
Q*BERT performs similarly to COMET-A2C and is likely due to inferring entities. 


The results for Experiment 2 are shown in Figure~\ref{fig:graphs} (right).
In this experiment agents must contend with missing object references in room descriptions.
KG-A2C never makes it past a score of 2---it enters the bathroom but cannot complete any tasks due to the inability to directly observe the sink, toilet, or shower.
KG-A2C-BERT performs identical to KG-A2C; BERT's commonsense procedural guidance doesn't help if it cannot generate the necessary commands to begin with.
COMET-A2C and Q*BERT are able to use the sink, toilet, and shower to successfully complete all the tasks required in the bathroom which leads to greater reward.
Q*BERT converges faster in this extreme experimental condition.
%
Experiment 2 confirms our intuitions about the role that commonsense inferences are playing in the agent's decision-making. 
By making the presence of key objects in one location implicit instead of explicit, we can verify in a controlled fashion that commonsense inferences are able to augment agents' senses. Similar to Experiment 1, Q*BERT more consistently reaches the end of the bathroom task. While COMET-A2C does reach the successfully complete the task, it requires more iterations to actually successfully complete the task. This is due to the difference in the way COMET-A2C and Q*BERT infer commonsense information to the knowledge graph. 
Both infer the existence of the missing entities, allowing them to progress through the game.
However, Q*BERT infers a diverse set of information using a small set of questions to be answered whereas COMET-A2C focuses on {\em HasA} relations.
This variance allows Q*BERT to solve the intermediate steps that are required to reach the end of the bathroom task by capturing information beyond what the {\em HasA} relation produces, such as using the question ``What attributes does X possess?'' that explicitly connects entities in richer ways.

The {\em 9:05} game is a ``slice of life game'', which requires the player to recreate behavior that might be also conducted routinely in the real-world. 
Slice of life games are extreme in their invocation of common locations and common situations; agents that can effectively use commonsense knowledge are going to naturally fare better than those without. 
Most text-based games have a mix of fantasy and science fiction elements along with common locations and situations and it remains to be seen if commonsense can help in these scenarios.


\section{Conclusions}

We conducted experiments in slice of life text-based games to understand how commonsense knowledge can help agents handle puzzles that involve locations and scenarios commonly found in the real world. 
Slice of life games, and {\em 9:05} in particular, are extreme versions of text-based games that require the player to recreate behavior that might be also conducted routinely in the real-world.
Although slice of life games are dominated by these scenarios, most text-based games interleave mundane world interactions with the world between fantasy and science fiction puzzles. 
Our experiments show that commonsense inferences can be used to augment an agent's beliefs about the state of the world, making the agent more robust against observation failures or against missing information in text descriptions.
We find that---regardless of the source of commonsense knowledge---augmenting the agent's world state beliefs is more successful than biasing the agent's exploration.


We contend that text-based, slice of life games are stepping stones toward goal-based natural language interactions with humans; situations in which an agent primarily understands the dynamic world through listening and acts to change the world by speaking.
It is natural for commonsense details to be omitted in such environments.
Our work shows how a deep reinforcement learning framework for ``acting through language'' can be made more robust to real-world natural language phenomena.


\bibliography{wordplay}

\begin{thebibliography}{10}

\bibitem{Bosselut2019COMETCT}
Antoine Bosselut, Hannah Rashkin, Maarten Sap, Chaitanya Malaviya,
  A.~Çelikyilmaz, and Yejin Choi.
\newblock Comet: Commonsense transformers for automatic knowledge graph
  construction.
\newblock In {\em ACL}, 2019.

\bibitem{devlin18}
Jacob Devlin, Ming{-}Wei Chang, Kenton Lee, and Kristina Toutanova.
\newblock {BERT:} pre-training of deep bidirectional transformers for language
  understanding.
\newblock {\em CoRR}, abs/1810.04805, 2018.

\bibitem{jericho}
Matthew Hausknecht, Prithviraj Ammanabrolu, Marc-Alexandre C{\^{o}}t{\'{e}},
  and Xingdi Yuan.
\newblock Interactive fiction games: A colossal adventure.
\newblock In {\em Thirty-Fourth AAAI Conference on Artificial Intelligence
  (AAAI)}, 2020.

\bibitem{narasimhan15}
Karthik Narasimhan, Tejas~D. Kulkarni, and Regina Barzilay.
\newblock Language understanding for text-based games using deep reinforcement
  learning.
\newblock In {\em EMNLP}, pages 1--11, 2015.

\bibitem{fulda17}
Nancy Fulda, Daniel Ricks, Ben Murdoch, and David Wingate.
\newblock What can you do with a rock? affordance extraction via word
  embeddings.
\newblock In {\em IJCAI}, pages 1039--1045, 2017.

\bibitem{zahavy18}
Tom Zahavy, Matan Haroush, Nadav Merlis, Daniel~J Mankowitz, and Shie Mannor.
\newblock Learn what not to learn: Action elimination with deep reinforcement
  learning.
\newblock In S.~Bengio, H.~Wallach, H.~Larochelle, K.~Grauman, N.~Cesa-Bianchi,
  and R.~Garnett, editors, {\em Advances in Neural Information Processing
  Systems 31}, pages 3562--3573. Curran Associates, Inc., 2018.

\bibitem{ammanabrolu}
Prithviraj Ammanabrolu and Mark~O. Riedl.
\newblock Playing text-adventure games with graph-based deep reinforcement
  learning.
\newblock In {\em Proceedings of 2019 Annual Conference of the North American
  Chapter of the Association for Computational Linguistics: Human Language
  Technologies, NAACL-HLT 2019}, 2019.

\bibitem{adolphs2019ledeepchef}
Leonard Adolphs and Thomas Hofmann.
\newblock Ledeepchef: Deep reinforcement learning agent for families of
  text-based games.
\newblock {\em arXiv preprint arXiv:1909.01646}, 2019.

\bibitem{jonmay}
Xusen Yin and Jonathan May.
\newblock Comprehensible context-driven text game playing.
\newblock {\em CoRR}, abs/1905.02265, 2019.

\bibitem{Ammanabrolu2020Graph}
Prithviraj Ammanabrolu and Matthew Hausknecht.
\newblock Graph constrained reinforcement learning for natural language action
  spaces.
\newblock In {\em International Conference on Learning Representations}, 2020.

\bibitem{Ammanabrolu2020HowTA}
Prithviraj Ammanabrolu, Ethan Tien, Matthew~J. Hausknecht, and M.~Riedl.
\newblock How to avoid being eaten by a grue: Structured exploration strategies
  for textual worlds.
\newblock {\em ArXiv}, abs/2006.07409, 2020.

\bibitem{Murugesan2020EnhancingTR}
K.~Murugesan, Mattia Atzeni, Pushkar Shukla, Mrinmaya Sachan, Pavan
  Kapanipathi, and Kartik Talamadupula.
\newblock Enhancing text-based reinforcement learning agents with commonsense
  knowledge.
\newblock {\em ArXiv}, abs/2005.00811, 2020.

\bibitem{Zhou2020EvaluatingCI}
Xuhui Zhou, Y.~Zhang, Leyang Cui, and Dandan Huang.
\newblock Evaluating commonsense in pre-trained language models.
\newblock In {\em AAAI}, 2020.

\bibitem{Lan2020ALBERT:}
Zhenzhong Lan, Mingda Chen, Sebastian Goodman, Kevin Gimpel, Piyush Sharma, and
  Radu Soricut.
\newblock Albert: A lite bert for self-supervised learning of language
  representations.
\newblock In {\em International Conference on Learning Representations}, 2020.

\bibitem{Speer2016ConceptNet}
Robyn Speer, Joshua Chin, and Catherine Havasi.
\newblock {ConceptNet} 5.5: An open multilingual graph of general knowledge.
\newblock In {\em AAAI}, 2016.

\bibitem{griffith2013policy}
Shane Griffith, Kaushik Subramanian, Jonathan Scholz, Charles~L Isbell, and
  Andrea~L Thomaz.
\newblock Policy shaping: Integrating human feedback with reinforcement
  learning.
\newblock In {\em Advances in neural information processing systems}, pages
  2625--2633, 2013.

\end{thebibliography}
\bibliographystyle{unsrt}



\end{document}